\def\year{2021}\relax
\pdfoutput=1
\documentclass[letterpaper]{article} 
\usepackage{aaai21}  
\usepackage{times}  
\usepackage{helvet} 
\usepackage{courier}  
\usepackage[hyphens]{url}  
\usepackage{graphicx} 
\usepackage{natbib}  
\usepackage{booktabs}
\usepackage{caption} 
\usepackage{multirow}
\usepackage{amsmath}
\usepackage[switch]{lineno}

\usepackage{amssymb}
\usepackage{tablefootnote}

\usepackage{color}

\usepackage{graphicx}
\usepackage{floatrow}
\newfloatcommand{capbtabbox}{table}[][\FBwidth]
\newfloatcommand{figurebox}{figure}[\nocapbeside][\dimexpr(\textwidth-\columnsep)/2\relax]
\newfloatcommand{tablebox}{table}[\nocapbeside][\dimexpr(\textwidth-\columnsep)/2\relax]

\usepackage{amsfonts}
\usepackage{amsthm}
\usepackage{pifont}%
\newcommand{\cmark}{\ding{51}}%
\title{Sparse Single Sweep LiDAR Point Cloud Segmentation via Learning Contextual Shape Priors from Scene Completion}

\author {
	Xu Yan\textsuperscript{\rm 1,2$\dagger$}, 
	Jiantao Gao\textsuperscript{\rm 2,4 $\dagger$}, 
    Jie Li\textsuperscript{\rm 1,3},
	Ruimao Zhang\textsuperscript{\rm 1,2},  
	Zhen Li\textsuperscript{\rm 1,2} \thanks{Corresponding author. $^\dagger$ Equal first authorship.}, \\
	Rui Huang\textsuperscript{\rm 1,3}, 
	Shuguang Cui\textsuperscript{\rm 1,2}
	 \\
}
\affiliations {
     \textsuperscript{\rm 1} The Chinese University of Hong Kong (Shenzhen), 
     \textsuperscript{\rm 2} Shenzhen Research Institute of Big Data,\\
     \textsuperscript{\rm 3} Shenzhen Institute of Artificial Intelligence and Robotics for Society,
	\textsuperscript{\rm 4} Shanghai University\\
	\textit{{\{xuyan1@link., {lizhen@}\}cuhk.edu.cn}}
	}

\begin{document}
	\maketitle
	\begin{abstract}
    LiDAR point cloud analysis is a core task for 3D computer vision, especially for autonomous driving. 
	However, due to the severe sparsity and noise interference in the single sweep LiDAR point cloud, the accurate semantic segmentation is non-trivial to achieve. 
	In this paper, we propose a novel sparse LiDAR point cloud semantic segmentation framework assisted by learned contextual shape priors.
	In practice, an initial semantic segmentation (SS) of a single sweep point cloud can be achieved by any appealing network and then flows into the semantic scene completion (SSC) module as the input.
    By merging multiple frames in the LiDAR sequence as supervision, the optimized SSC module has learned the contextual shape priors from sequential LiDAR data, completing the sparse single sweep point cloud to the dense one.
    Thus, it inherently improves SS optimization through fully end-to-end training.
    Besides, a Point-Voxel Interaction (PVI) module is proposed to further enhance the knowledge fusion between SS and SSC tasks, i.e., promoting the interaction of incomplete local geometry of point cloud and complete voxel-wise global structure.
    Furthermore, the auxiliary SSC and PVI modules can be discarded during inference without extra burden for SS.
	Extensive experiments confirm that our JS3C-Net achieves superior performance on both SemanticKITTI and SemanticPOSS benchmarks, i.e., 4\% and 3\% improvement correspondingly.

	\end{abstract}
	
	\section{Introduction}

	
	
    LiDAR point clouds, compared with data from other sensors, such as cameras and radars in autonomous driving perception, have advantages of both accurate distance measurements and fine semantic descriptions.
    Semantic segmentation of LiDAR point clouds is usually conducted by assigning a semantic class label to each point. It is traditionally viewed as a typical task in the computer vision community. 
    In autonomous driving, accurate and effective point cloud semantic segmentation undoubtedly plays a critical role.

	\begin{figure}[t]
		
	\noindent\includegraphics[width=0.95\columnwidth, height=4.5cm]{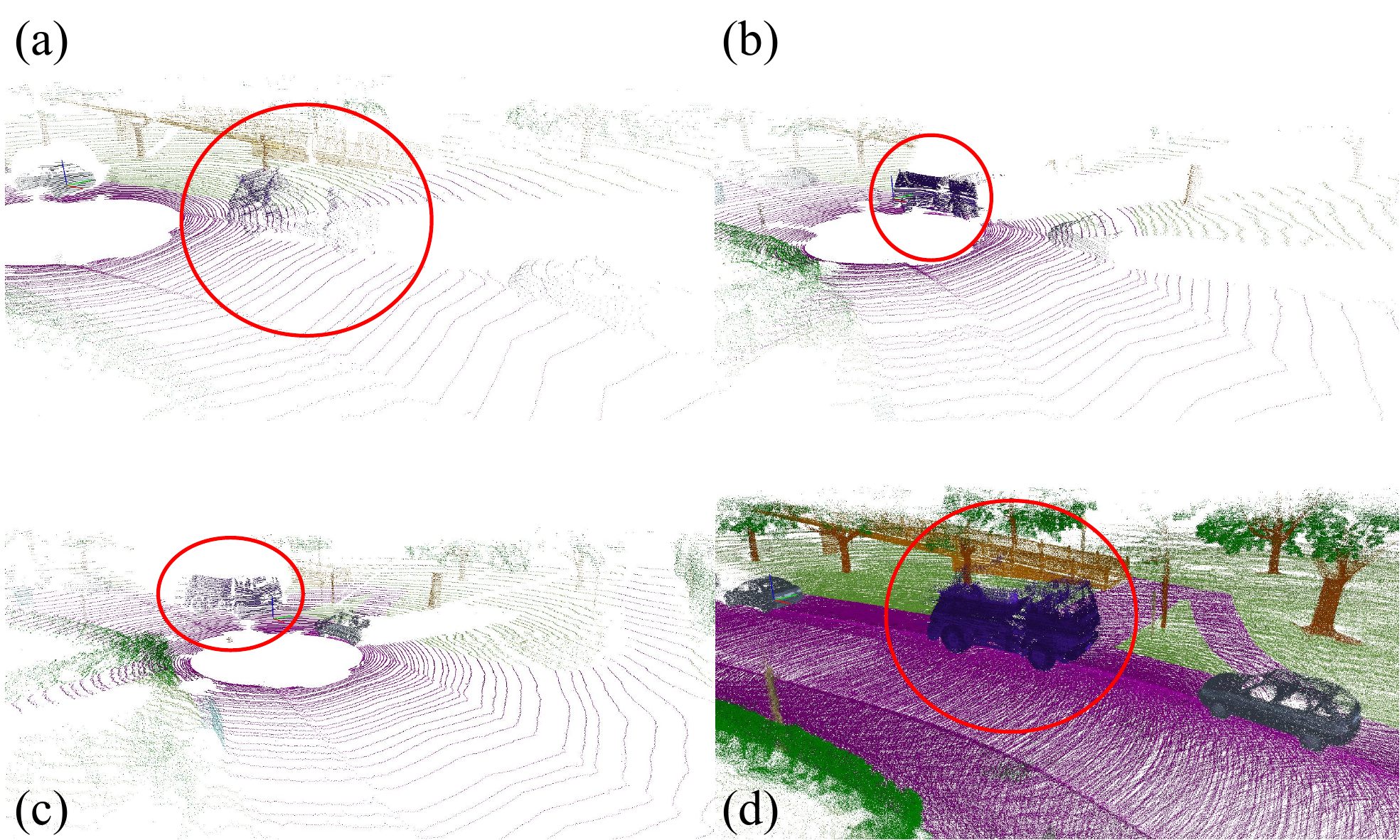}
		
		\caption{{\bf Learning shape priors from multiple frames.} 
		For the sparse per-sweep point cloud shown in (a), it is nontrivial for current methods to recognize the truck from partial components. 
		However, if we introduce the auxiliary information from adjacent frames (b) and (c), it is much easier to segment the complete truck in (d).}
		\label{fig:fig1}
		
	\end{figure}

	%
	Previous studies~\cite{Thomas_2019_ICCV,PointConv} about point cloud semantic segmentation mainly focused on the complete or dense point cloud scenarios, which are post-processed by merging multiple collected LiDAR or RGB-D sequences (e.g., ScanNet~\cite{scannet}, S3DIS~\cite{s3dis} and Semantic3D~\cite{hackel2017semantic3d}). 
	However, raw per-sweep LiDAR point clouds, as the original input of autonomous driving, are much sparser. 
	Their sparsity usually increases with the reflection distance, which often leads to extremely shapes missing and uneven point sampling for various categories. 
	Therefore, despite the promising performance on complete data (e.g., 80\% mIOU on Semantic3D), the semantic segmentation of sparse single sweep LiDAR point cloud still remains a big challenge, which extremely limits its accuracy in real applications.
	%

    In this paper, we try to break through the barrier of semantic segmentation on sparse single sweep LiDAR point clouds. 
    One plausible way to solve this problem is to fully utilize the sequential nature of LiDAR data. 
    Taking the scenario shown in Fig.~\ref{fig:fig1}(a) as an example, for a per-sweep point cloud with extremely sparse points of the truck, it seems impossible for previous methods to conduct accurate segmentation. Nevertheless, such segmentation would be possible, if we introduce the richer shape information from the other two frames, i.e., Fig.~\ref{fig:fig1}(b) and Fig.~\ref{fig:fig1}(c), to reconstruct a shape-complete truck as shown in Fig.~\ref{fig:fig1}(d).
	For this purpose, some previous works utilized historical adjacent frames to supplement the local details missing from the point clouds. 
	For instance, SpSequenceNet~\cite{shi2020spsequencenet} and MeteorNet~\cite{liu2019meteornet} use the point cloud of the current frame to query the nearest neighbors from the previous frames, following which a feature aggregation is conducted to fuse the adjacent-frame information.
	PointRNN~\cite{fan2019pointrnn} applies Recurrent Neural Networks (RNNs) to select available features from previous scenes.
	However, all of the above methods become unavailable in most real scenarios since the following reasons: 
	%
	(1) These methods exclusively use historical frames of the current scene in LiDAR sequence. 
	Thus, they cannot introduce priors for newly incoming objects in this scene, i.e., they cannot utilize future frames. 
	(2) Their proposed feature aggregation methods (i.e., through kNN or RNN) inevitably increase the computational burden, which makes it less effective and unsuitable for self-driving task.
	

	To solve the above issues, we propose an enhanced \textbf{J}oint single sweep LiDAR point cloud \textbf{S}emantic \textbf{S}egmentation by exploiting learned shape prior form \textbf{S}cene \textbf{C}ompletion network, i.e., \textbf{JS3C-Net}.
	%
	Specifically, by merging dozens of consecutive frames in a LiDAR sequence, a large complete point cloud is achieved as ground truth for the Semantic Scene Completion (SSC) task without extra annotation.
	%
    The optimized SSC by using these annotations could capture the compelling shape priors, making the incomplete input complete to the acceptable shape with semantic labels~\cite{Song2017Semantic}.
	Therefore, the completed shape priors can inherently benefit the semantic segmentation (SS) due to the fully end-to-end training strategy. 
    Furthermore, a well-designed Point-Voxel Interaction (PVI) module is further proposed for implicit mutual knowledge fusion between the SS and SSC tasks.
	Concretely, the point-wise segmentation and voxel-wise completion are leveraged to maintain the coarse global structure and fine-grained local geometry through PVI module.
	%
	%
	%
    %
	More importantly, we design our SSC and PVI modules to be \textbf{disposable}.
	To achieve this, JS3C-Net combines the SS and SSC in a cascaded manner, which means that it would not influence the information flow for SS while discarding the SSC and PVI modules in inference stage. 
	Thus, it can prevent bringing the extra computing burden from generating complete high-resolution dense volumes.
	%
	%

	Our main contributions are: 
	1) To the best of our knowledge, the proposed \textbf{JS3C-Net} is the first to achieve the enhanced sparse single sweep LiDAR semantic segmentation via auxiliary scene completion.
	2) For better trade-off between performance and effectiveness, our auxiliary components are designed in cascaded and disposable manners, and a novel point-voxel interaction (PVI) module is proposed for better feature interaction and fusion between the two tasks. 
	3) Our method shows superior results in both SS and SSC on two benchmarks, i.e., SemanticKITTI~\cite{behley2019semantickitti} and SemanticPOSS~\cite{pan2020semanticposs}, by a large margin.
	
	\section{Related Work}
	\subsubsection{Point Cloud Semantic Segmentation.} Unlike 2D images with regular grids, point clouds are often sparse and disordered. Thus, point clouds processing is a challenging task. There are three main strategies to approach this problem: \textbf{projection-based}, \textbf{voxel-based} and \textbf{point-based}. 
	(1) Projection-based methods map point clouds onto 2D pixels, so that traditional CNN can play a normal role. Previous works projected all points scanned by the rotating LiDAR onto 2D images by plane projection~\cite{lawin2017deep, boulch2017unstructured, tatarchenko2018tangent} or spherical projection~\cite{wu2018squeezeseg,wu2019squeezesegv2}. 
	(2) Considering the sparsity of point clouds and memory consumption, it is not very effective to directly voxelize point clouds and then use 3D convolution for feature learning. Various subsequent improved methods have been proposed, e.g.,  efficient spatial sparse convolution~\cite{choy20194d, SparseConv} and octree based convolutional neural networks~\cite{wang2017cnn,riegler2017octnet}. Also \cite{tang2020searching} use NAS to obtain a more efficient feature representation. 
	(3) Point-based methods directly process raw point clouds~\cite{qi2017pointnet,qi2017pointnet++}.  Usually, most methods use sampling strategies to select sub-points from the original point clouds, and then use local grouping with feature aggregation function for local feature learning of each sub-point. Among these methods, graph-based
	learning~\cite{wang2019graph,landrieu2018large,landrieu2019point,DGCNN} and convolution-like operations~\cite{Thomas_2019_ICCV,PointConv,hu2019randla} are widely used. However, previous methods often suffer from local information missing in LiDAR scenes due to insufficient priors and bias in data collection.
	
	\begin{figure*}[t]
		
		\noindent\includegraphics[width=0.95\textwidth]{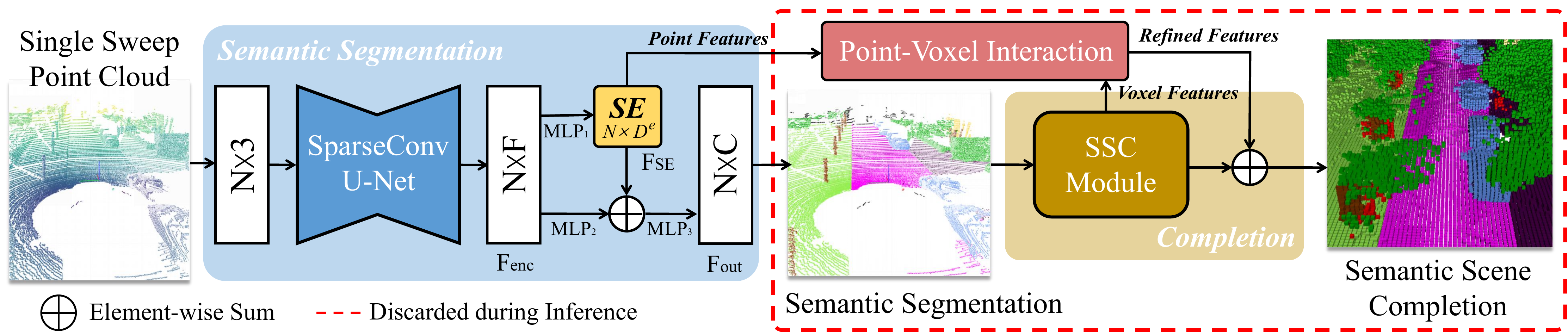}
		
		\caption{{\bf Overall pipeline of JS3C-Net}. Given a sparse incomplete single sweep point cloud, it firstly uses a sparse convolution U-Net to conduct point feature encoding $F_{enc}$. 
		Based on the initial encoding, MLP${_1}$ is used to generate shape embedding (SE) $F_{SE}$, which flows into MLP${_3}$ together with initial encoding transferred through MLP${_2}$ to generate $F_{out}$ for point cloud semantic segmentation.
		Afterwards, the incomplete fine-grained point features from SE and complete voxel features from semantic scene completion (SSC) module flows into the Point-Voxel Interaction (PVI) module to achieve the refined features, which finally outputs the completion voxels with supervision.
		Note that the SSC and PVI modules can be discarded during inference. 
	}
		%
		
		\label{fig:fig2}
		
	\end{figure*}
	
	\subsubsection{Semantic Scene Completion.}
	Semantic scene completion (SSC) aims to produce a complete 3D voxel representation from an incomplete input. Concretely, Song \textit{et al}.\cite{Song2017Semantic} firstly use single-view depth as input to construct an end-to-end model SSCNet, which can predict the results of scene completion and semantic labeling simultaneously. Spatially sparse group convolution is used in Zhang \textit{et al}.~\cite{Jiahui2019Efficient} for fast 3D dense prediction. 
	Meanwhile, coarse-to-fine strategies (e.g., LSTM based model) are used in~\cite{Dai2017ScanComplete,Han2017High} to recover missing parts of 3D shapes.
	More recently, some works introduce color information~\cite{Garbade_2019_CVPR_Workshops} and use more powerful two-stream feature extractor~\cite{li2019rgbd} or feature fusion~\cite{liu20203d} to enhance the performance. However, SSC is rarely studied in large-scale LiDAR scenarios, and the serious geometric details missing and real-time requirements make it difficult.
	
	
	\subsubsection{Multi-task Learning on Segmentation.}
	Multi-task learning aims to improve the learning efficiency and prediction accuracy for each task through knowledge transfer, which is widely used in 2D images segmentation~\cite{kendall2018multi}. Wang \textit{et al}.~\cite{wang2019associatively}, Pham \textit{et al}.~\cite{pham2019jsis3d} and Wei \textit{et al}.~\cite{wei2020multi} innovatively combine semantic and instance segmentation with specific-designed fusion modules to improve the performance. OccuSeg~\cite{han2020occuseg} proposes a 3D voxel projection-based segmentation network with voxel occupancy size regression and owns advantages of robustness in prediction. However, the shape priors brought by completion tasks are often neglected in previous works, while a proper use could improve the performance of segmentation.
	
	\section{Method}
	\subsection{Overview}
	
	The pipeline of {\textbf{JS3C-Net}\footnote{\url{https://github.com/yanx27/JS3C-Net}.}} is illustrated in Fig.~\ref{fig:fig2}. In practice, we firstly use the general appealing point cloud segmentation network to obtain initial point semantic segmentation and a shape embedding (SE) for each incomplete single-frame point cloud. 
	Then SSC module takes results of segmentation network as input and generates the completed voxel of the whole scene with dense convolution neural network.
	Meanwhile, a point-voxel interaction (PVI) module is proposed to conduct shape-aware knowledge transfer. 
	
	\subsection{Semantic Segmentation}
	
	In general, a point cloud has two components: the points $\mathcal{P} \in \mathbb{R}^{N \times 3}$ and their features $\mathcal{F} \in \mathbb{R}^{N \times D}$, where the points record spatial coordinates of $N$ points and $D$-dimensional features can include any point-wise information, e.g., RGB information. 
	Here we only use point coordinates as inputs.
	%
	
	For semantic segmentation stage, we simply choose Submanifold SparseConv~\cite{SparseConv} as our backbone.
	Unlike traditional voxel-based methods~\cite{ronneberger2015u,choy20194d} directly transforming all points into the 3D voxel grids by averaging all input features, it only stores non-empty voxels by the Hash table, and conduct convolution operations only on these non-empty voxels with more efficient way. 
	Afterwards, the voxel-based output from sparse convolution based U-Net (SparseConv U-Net) is transformed back to the point-wise features $\mathcal{F}_{enc} \in \mathbb{R}^{N \times F}$ by nearest-neighbor interpolation. 
	To further introduce shape priors (see latter section) to point-wise features, we use multi-layer perceptions (MLP${_1}$) to transfer their features to shape embedding (SE) $F_{SE} \in \mathbb{R}^{N \times D^e}$, which works as the input of the subsequent point-voxel interaction module. 
	Furthermore, an element-wise addition operation after MLP${_2}$ is used to fuse shape embedding with features from SparseConv U-Net. 
	Finally, $F_{out} \in \mathbb{R}^{N \times C}$ are generated by MLP${_3}$ and prepare for further semantic scene completion stage.
	
	\begin{figure*}[t]
		
		\noindent\includegraphics[width=0.95\textwidth,height=5.5cm]{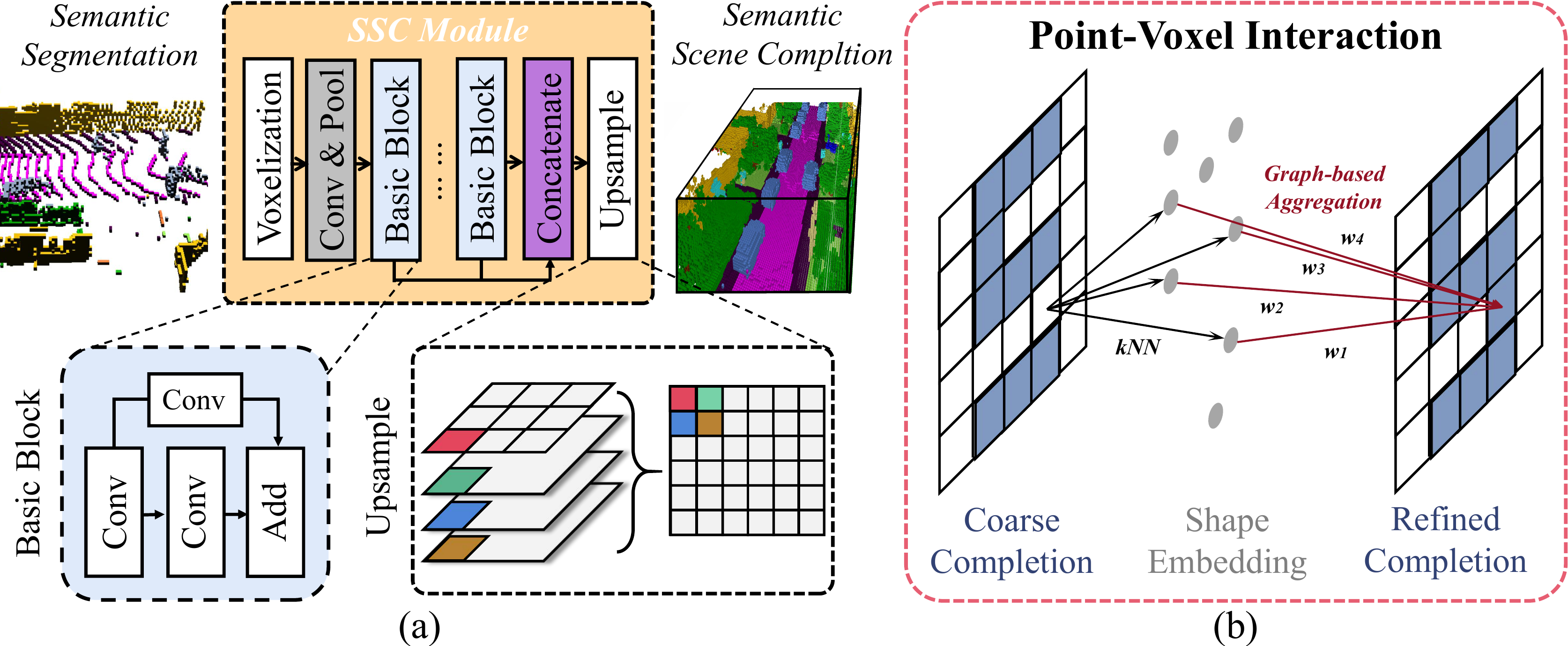}
		
		\caption{Part (a) shows the inner structure of SSC module, which uses semantic probability from segmentation network as inputs, generating complete volume by several convolution blocks and dense upsample. Part (b) illustrates a 2D case of PVI module, which uses the center points of the coarse global structure of number '5' to query $k$ nearest neighbors from the raw point cloud, and then applies graph-based aggregation to achieve the completed '5' through fine-grained local geometry.}
		\label{fig:fig3}
		
	\end{figure*}
	
	\subsection{Cascaded Semantic Scene Completion}
	\label{completion}
    The Semantic Scene Completion (SSC) module aims to introduce contextual shape priors from the entire LiDAR sequence. For stage of SSC, it takes the semantic probability $F_{out}$ from the SparseConv as input, and then predict the completion results.
	
	The architecture of our SSC module is depicted in Fig.~\ref{fig:fig3} (a). 
	Taking an incomplete point cloud with per points categorical probability as input, the network firstly conducts voxelization to obtain high-resolution 3D volume, and uses one convolution layer following by a pooling layer to reduce the resolution and the complexity of computation. 
	Then, several basic blocks using convolutions with skip-connection are exploited to learn a local geometry representation. 
	Afterwards, the features from different scales are concatenated to aggregate information from multiple scales. 
	For achieving original resolution of SSC output, we leverage dense upsampling~\cite{liu2018see} shown in Fig.~\ref{fig:fig3}(a) to avoid the interpolation inaccuracy, instead of dilation convolution based 
	upsampling ~\cite{Song2017Semantic}.
	%
	%
	Finally, we obtain a voxel output with $C+1$ channels ($C$ semantic categories label and one non-object label).
	This coarse completion will be fed into the PVI module for further mutual enhancements.

	\subsection{Shape-aware Point-Voxel Interaction}
	
	\label{interaction}
	To fully utilize implicit knowledge transfer for mutual improvements of two tasks, we innovatively propose a shape-aware Point-Voxel Interaction (PVI) module for knowledge fusion between incomplete point clouds and complete voxels from former two steps. 
	Although the SSC module can generate voxel-wise output with complete shape, such output is relatively coarse due to the voxelization procedure, leading to local geometric details missing. 
	The entire geometric representation in raw point cloud data can, nevertheless, provide semantic guidance during completion process despite missing parts. 
	
	The inner structure of PVI module is shown in Fig.~\ref{fig:fig3} (b), which aims to conduct a coarse-to-fine process for the SSC prediction.
	To be more precise, per point shape embedding $\mathcal{F}_{SE} \in \mathbb{R}^{N\times D^e}$ and coarse completion from SSC module $\mathcal{V}$ flow into PVI module as input.
	Afterwards, PVI firstly selects geometric centers of all non-empty voxels from $\mathcal{V}$ as a new point cloud $\mathcal{P}^v \in \mathbb{R}^{N' \times(C+1)}$, then it uses k-nearest neighbor by Euclidean distance to query the closest points from original point cloud $\mathcal{P}$. 
	To this end, a graph convolutional network is further employed to enhance the relation learning between $\mathcal{P}^v$ and $\mathcal{P}$ in both spatial and semantic spaces. 
	In particular, the nodes of the graph are defined by the point positions with associated points features.
    For each point $p^v_i \in \mathcal{P}^v$ and its $j$-th neighboring point $p_j \in \mathcal{P}$, we adopt the convolutional operator from DGCNN~\cite{DGCNN} to define edge-features ${e}_{ij}$ between two points as:
	\begin{equation}
	{e}_{ij} = \phi([p^v_i,f^v_i], [p^v_i,f^v_i]-[p_j,f_j]),
	\end{equation}
	where $f^v_i$ and $f_j$ are features of point $p^v_i$ and $p_j$ respectively, and $[\cdot,\cdot]$ means concatenation operation.
	The share-weighted non-linear fuction $\phi$ is the multi-layer perceptron (MLP) in this paper (any differentiable architecture alternative).
	Finally, by stack $l$ graph convolutional network (GCN) layers, we obtain final fine-grained completion.

    The feature interaction process enables features of sparse point cloud to acquire the ability to predict semantics of complete voxels. 
    Therefore, the information on complete details can positively affect the segmentation part through back propagation.
	Furthermore, PVI module enhances the probability of predicting whether the corresponding voxel of $p^v_i$ represents an object in the fine-grained architecture, which fully utilizes spatial and semantic relationships between each $p_j$ and $p^v_i$. 
	Finally, this enhanced feature will be added to the original coarse completion output through a residual connection for further refinement (see the refined number '5' in Fig.~\ref{fig:fig3} (b)).

	\subsection{Uncertainty-weighted Multi-task Loss}
	In order to further balance these two tasks and avoid complicated manual attempts during the end-to-end training, we use the uncertainty weighting method proposed in \cite{kendall2018multi}. It introduces acquirable parameters to automatically adjust the optimal proportion between different tasks. Specifically, the joint loss can be written as:
	\begin{equation}
	\begin{aligned}
	\label{eq2}
	\mathcal{L}(W, \sigma_1, \sigma_2) = \frac{1}{2\sigma_1^2}\mathcal{L}_{seg}(W_{seg})+\frac{1}{2\sigma_2^2}\mathcal{L}_{complet}(W)\\
	+log \sigma_1 + log \sigma_2,
	\end{aligned}
	\end{equation}
	where losses $\mathcal{L}_{seg}$ for segmentation and $\mathcal{L}_{complet}$ for completion are both weighted cross-entropy losses exploited to update the network parameters $W$. Note that gradients from segmentation outputs are only conducted on parameters of segmentation network $W_{seg}\in W$. In addition, we use trainable parameters $\sigma_1$ and $\sigma_2$ to weight the proportion of these two tasks for optimal trade-off. Their uncertainty can be deduced as two $log$ term to control their values. Finally, during the training process, these two tasks will promote each other through back-propagation of joint learning.

	\begin{table*}[!htp]

	\renewcommand\tabcolsep{1pt} 
	
	\begin{center}
		\resizebox{\textwidth}{!}{
		\begin{tabular}{lcc|ccccccccccccccccccc}
			\hline
			
			Method& Size&
							\rotatebox{90}{mIoU}&
							\rotatebox{90}{road}&
							\rotatebox{90}{sidewalk}&
							\rotatebox{90}{parking}&
							\rotatebox{90}{other-ground}&
							\rotatebox{90}{building}&
							\rotatebox{90}{car}&
							\rotatebox{90}{truck}&
							\rotatebox{90}{bicycle}&
							\rotatebox{90}{motorcycle}&
							\rotatebox{90}{other-vehicle}&
							\rotatebox{90}{vegetation}&
							\rotatebox{90}{trunk}&
							\rotatebox{90}{terrain}&
							\rotatebox{90}{person}&
							\rotatebox{90}{bicyclist}&
							\rotatebox{90}{motorcyclist}&
							\rotatebox{90}{fence}&
							\rotatebox{90}{pole}&
							\rotatebox{90}{traffic sign} \\
										
			\hline
			\hline
			
			SqueezeSegV2~\cite{wu2019squeezesegv2}&\multirow{5}{1.5cm}{\centering64*2048 pixels}  &39.7&88.6& 67.6& 45.8& 17.7& 73.7& 81.8& 13.4& 18.5& 17.9& 14.0& 71.8& 35.8 &60.2& 20.1& 25.1& 3.9& 41.1& 20.2& 26.3\\
			
			DarkNet53Seg~\cite{behley2019semantickitti}& &49.9 &\bf{91.8} &74.6 &64.8 &27.9 &84.1 &86.4 &25.5 &24.5 &32.7 &22.6 &78.3 &50.1 &64.0 &36.2 &33.6 &4.7 &55.0 &38.9 &52.2\\
			
			RangeNet53++~\cite{milioto2019rangenet++}& &52.2 &\bf{91.8} &\bf{75.2} &\bf{65.0} &27.8 &87.4 &91.4 &25.7 &25.7 &34.4 &23.0 &80.5 &55.1 &64.6 &38.3 &38.8 &4.8 &58.6 &47.9 &55.9\\
			
			3D-MiniNet~\cite{alonso20203d}& &55.8 &91.6 &74.5 &64.2 &25.4 &89.4 &90.5 &28.5 &42.3 &42.1 &29.4 &82.8 &60.8 &66.7 &47.8 &44.1 &14.5 &60.8 &48.0 &56.6\\
			SqueezeSegV3~\cite{xu2020squeezesegv3}& &55.9 &91.7 &74.8 &63.4 &26.4 &89.0 &92.5 &29.6 &38.7 &36.5 &33.0 &82.0 &58.7 &65.4 &45.6 &46.2 &20.1 &59.4 &49.6 &58.9\\
			
			\hline
			
			PointNet++~\cite{qi2017pointnet++}& \multirow{5}{1.5cm}{\centering50K pts}&20.1& 72.0 &41.8 &18.7& 5.6 &62.3 &53.7 &0.9 &1.9 &0.2& 0.2& 46.5 &13.8 &30.0 &0.9 &1.0 &0.0 &16.9 &6.0 &8.9\\
			TangentConv~\cite{tatarchenko2018tangent}& &40.9 &83.9 &63.9 &{33.4} &{15.4} &{83.4} &{90.8} &15.2&{ 2.7}& 16.5 &12.1 &79.5 &49.3 &58.1 &23.0 &28.4 &{8.1} &{49.0} &35.8 &28.5\\
			PointASNL~\cite{yan2020pointasnl} & & 46.8&87.4 &74.3&24.3&1.8&83.1&87.9&39.0&0.0&25.1&29.2&84.1&52.2&\bf{70.6}&34.2& 57.6&0.0&43.9&57.8&36.9\\
			RandLA-Net~\cite{hu2019randla}& &55.9 &90.5 &74.0 &61.8 &24.5 &89.7 &94.2 &43.9 &29.8 &32.2 &{39.1} &83.8 &63.6 &68.6 &48.4 &47.4 &9.4 &60.4 &51.0 &50.7\\
			KPConv~\cite{Thomas_2019_ICCV}& &58.8 &{90.3} &72.7 &{61.3} &31.5 &90.5 &95.0 &33.4 &30.2 &42.5 &44.3 &84.8 &69.2 &69.1 &61.5 &61.6 &11.8 &64.2 &56.4 &47.4\\
			\hline
			
			PolarNet~\cite{zhang2020polarnet}& \multirow{3}{1.5cm}{\centering50K pts}&54.3 &90.8 &74.4 &61.7 &21.7 &90.0 &93.8 &22.9 &40.3 &30.1 &28.5 &84.0 &65.5 &67.8 &43.2 &40.2 &5.6 &61.3 &51.8 &57.5\\

			SparseConv (Baseline) &  &61.8 &89.9&72.1&56.5&29.6&90.5&{94.5}& 43.5 & {51.0}& {42.4}&31.3&83.9 & 67.4 &68.3& {60.4}&61.3& \bf{41.1} &65.6&57.9 & 67.7\\
			
			\textbf{JS3C-Net (Ours)} &  & \bf{66.0} &88.9&72.1&61.9&\bf{31.9}&\bf{92.5}&\bf{95.8}& {\underline{\bf{54.3}}}
      	&{\underline{\bf{59.3}}}&{\underline{\bf{52.9}}}&{\underline{\bf{46.0}}}&\bf{84.5} & \bf{69.8} &67.9&{\underline{\bf{69.5}}}&{\bf{65.4}}& {{39.9}} &\bf{70.8} &\bf{60.7} &{\bf{68.7}}\\
			\hline
			
		\end{tabular}
	}
	\end{center}
\caption{Semantic segmentation on the \textit{SemanticKITTI} benchmark. {{Underline}} marks results that $\sim 10\%$ higher than baseline.}
	\label{tab:seg}
\end{table*}


    \subsection{Disposable Properties of Auxiliary Components}
    
	Our proposed JS3C-Net is a general joint learning framework to improve the point cloud segmentation by introducing complete shape extracted by LiDAR sequence itself. 
	The network used in semantic segmentation stage is flexible and can be replaced by other appealing networks. 
	%
	Furthermore, our JS3C-Net is effective enough for real-time applications, since the auxiliary components (i.e., SSC module and PVI module) can be discarded during inference to prevent introducing any computing burden for segmentation.
	That is to say, the completion part are \textit{only} exploited in the training process as the dotted line shown in Fig.~\ref{fig:fig2}.
	
	\section{Experiments}
	
	\subsection{Dataset}
	
	To further verify the effectiveness of our method, we evaluate JS3C-Net on two benchmarks SemanticKITTI~\cite{behley2019semantickitti} and SemanticPOSS~\cite{pan2020semanticposs}.
	SemanticKITTI is currently the largest LiDAR sequential dataset with point-level annotations, which consists of $43 552$ densely annotated LiDAR scans belonging to $21$ sequences. These scans are annotated with a total of $19$ valid classes and each scan spans up to $160\times160\times20$ meters with more than $\sim 10^5$ points.
	We follow the official split for training, validation and online testing.
	SemanticPOSS is a newly proposed dataset with 11 similar annotated categories with SemanticKITTI.
	However, it is more challenging because each scene contains more than \textit{twenty times} sparse small objects (i.e., people and bicycle), while the total frames number are only 1/20 of SemanticKITTI.
	
    On both two datasets, we firstly merge consecutive 70 frames for every single frame to generate the complete volume of the entire scans. 
	Then we select a volume of $51.2m$ in front of the LiDAR, $25.6m$ to each side of the LiDAR, and $6.4m$ in height with the resolution of $0.2m$, which results in a volume of $256\times256\times32$ voxels for prediction. 
	Each voxel is assigned a single label based on a majority vote over all labeled points inside a voxel. Voxels containing no point are labeled as empty voxels. 
	Note that voxels absent in all frames will not be considered in the loss calculation and evaluation.

	\subsection{Joint Learning Protocol} 
	During the end-to-end training process of JS3C-Net, we use the Adam optimizer as our optimizer. 
	The batch size is set to 6 for the total 50 epochs. 
	The initial learning rate is set as $0.001$ and decreases by $30\%$ after every $5$ epochs. 
	The weighted terms $\sigma_1$ and $\sigma_2$ in Eqn.~\ref{eq2} are randomly initialized and are trained with $\times 10$ learning rates. 
	For semantic segmentation, $0.05m$ grids are used to conduct voxelization in SparseConv model. 
	We randomly rotate the input point cloud along the y-axis during the training process and randomly scale it in the range of $0.9$ to $1.1$. 
	During the inference, we apply the general voting strategy~\cite{Thomas_2019_ICCV,hu2019randla} to average multiple prediction results of randomly augmented point clouds.
	%
	%
	Similar data augmentation and voting strategies are also exploited for semantic scene completion. However, due to the particularity of the input volume, we use random flip along the x-axis and z-axis, and randomly rotate along y-axis by $90$ degrees. 
	%
	%
	All experiments are conducted on an Nvidia Tesla V100 GPU. 
    More network details will be described in the supplementary material.

	\begin{figure*}[t]
		
		\noindent\includegraphics[width=0.95\textwidth, height=7cm]{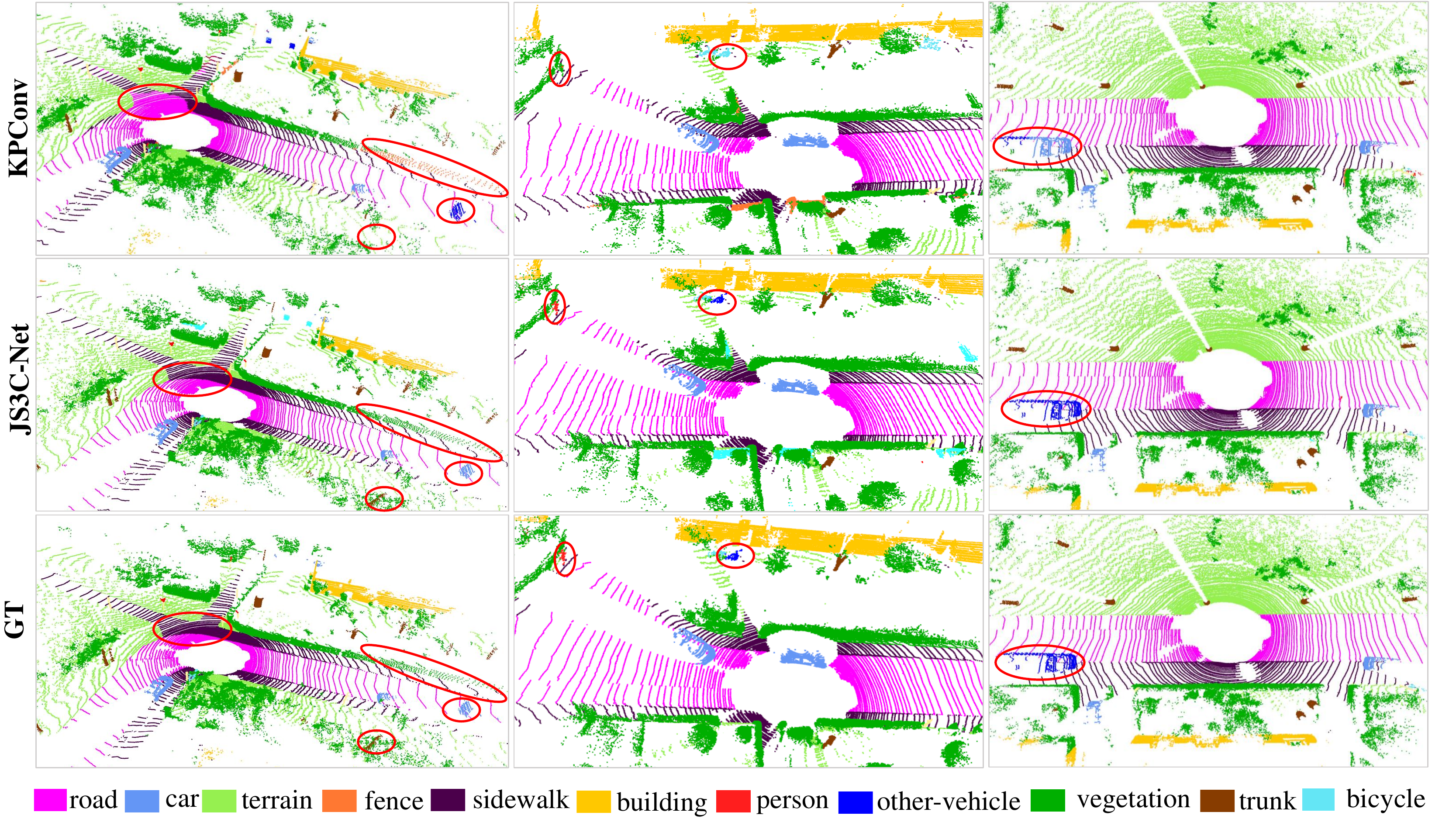}
		
		\caption{Qualitative results of JS3C-Net on the validation set of \textit{SemanticKITTI}~\cite{behley2019semantickitti}. {\color{red} Red} circles show that our method performs better in many details than recent state-of-the-art KPConv~\cite{Thomas_2019_ICCV}. Results for \textit{SemanticPOSS} dataset are illustrated in supplementary material.}
		\label{fig:fig4}
		
	\end{figure*}
	
\begin{table}
	\small
		\caption{Semantic segmentation results on the \textit{SemanticPOSS} benchmark. The upper, medium and bottom parts contain previous projection-based, point-based and voxel-based methods, respectively.
		}
		\begin{center}
		\begin{tabular}{l|ccc|c}
    		\hline
    		&
    		\multicolumn{3}{c|}{Selected 3 classes}
    		&11 classes 
    		\\
    		Method
    		&{People}
    		&{Rider}
    		&{Bike}
    		&{avg IoU} \\
    		\hline
    		\hline
    		SequeezeSegV2 & 18.4 &11.2&32.4 &29.8 \\\hline
        	PointNet++ & 20.8& 0.1& 0.1 &20.1 \\
    
    
    		RandLA-Net & 69.2 & 26.7 & 43.9 & 53.5 \\
    		KPConv & 77.3 & 29.4 & 53.2 &55.2 \\
    		\hline
    		SparseConv &76.3 &30.5 &53.5  &57.2\\
    		\textbf{JS3C-Net (Ours)} & \textbf{80.0} & \textbf{39.1} &\textbf{59.8} & \textbf{60.2}\\
    		\hline
	    \end{tabular}
		\end{center}
		
		\label{tab:seg_pose}
	\end{table}

	\subsection{Semantic Segmentation}
	
	In Tab.~\ref{tab:seg}, we compare our JS3C-Net with recent methods on SemanticKITTI benchmark.
	The upper, medium and bottom parts of the table contain projection-based, point-based and voxel-based methods, respectively. 
	The class averaged interactions over union (mIoU) is used in evaluation. 
	
	As shown in Tab.~\ref{tab:seg}, our JS3C-Net surpasses all existing methods by a large margin. 
	Merely using the SparseConv~\cite{graham2017submanifold}, training from scratch already improves upon prior arts.
	Yet, using our joint-learning strategy achieves markedly better segmentation results in mIoU.
	Specifically, JS3C-Net achieves significant improvements on small objects (e.g., motorcycle, bicycle and etc), where these objects always lose geometric details during the LiDAR collection. 
	Thanks to the contextual shape priors from SSC, our JS3C-Net
	can segment them well.
	Meanwhile, Fig.~\ref{fig:fig4} presents some visualization results of JS3C-Net on the validation split, which demonstrates great improvements on small objects, in particular.
	
	Tab.~\ref{tab:seg_pose} illustrates the semantic segmentation results on SemanticPOSS dataset, where we compare result state-of-the-art methods.
	These  results show that our proposed JS3C-Net can achieve larger improvement compared with our baseline on more challenging data with remarkable small objects.
	
	\begin{figure*}[t]
		\noindent\includegraphics[width=0.95\textwidth, height=5.9cm]{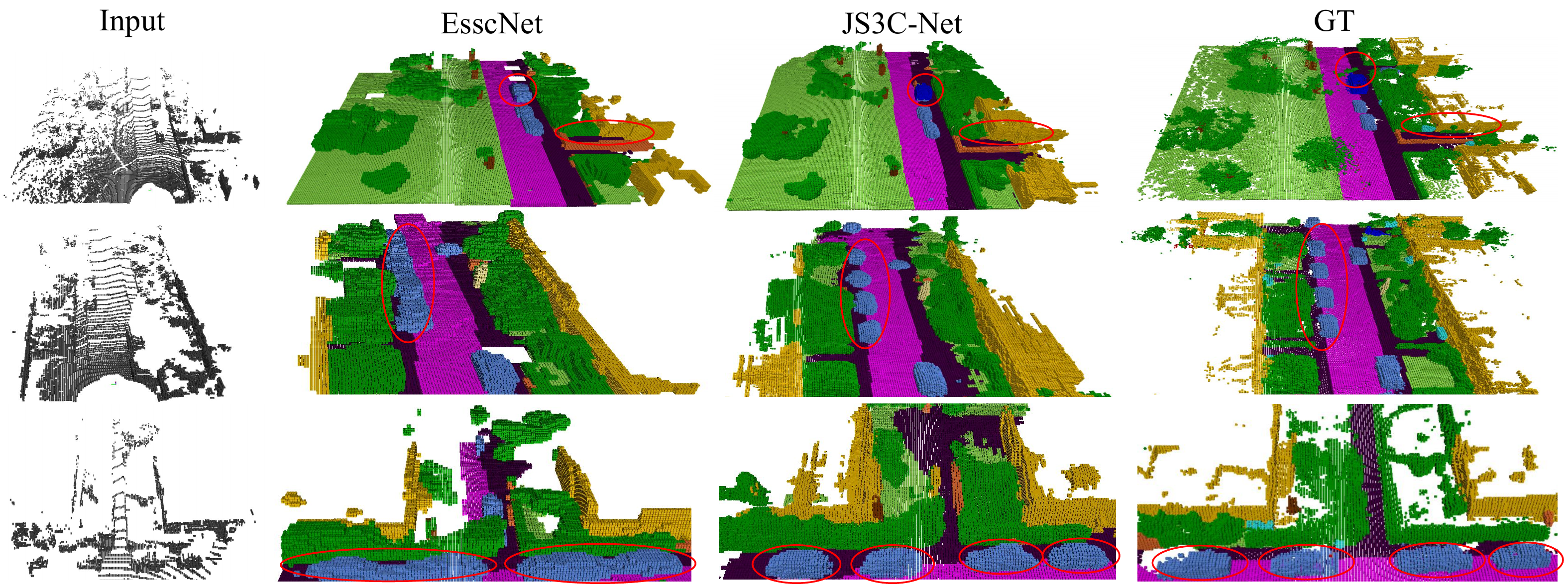}
		
		\caption{Qualitative results of SSC task on the validation set of \textit{SemanticKITTI}~\cite{behley2019semantickitti}.}
		\label{fig:fig5}
		
	\end{figure*}
	
	\begin{table}
	\small
		\caption{Semantic scene completion results on the \textit{SemanticKITTI} benchmark. Only the recent published approaches are compared.
		}
		\begin{center}
		\begin{tabular}{lccc|c}
    		\hline
    		Method
    		&{precision}&
    		{recall}&
    		{IoU}&	
    		{mIoU} \\
    		\hline
    		\hline
    		
        	SSCNet &31.7 &83.4 &29.8  &9.5\\
    		TS3D &31.6 &84.2 &29.8 &9.5\\
    		TS3D$^2$ &25.9 &\bf{88.3} &25.0 &10.2\\
    		EsscNet &62.6 &55.6 &41.8 &17.5\\
    		TS3D$^3$ &\bf{80.5} &57.7 &50.6 &17.7\\
    		\hline
    		\textbf{JS3C-Net (Ours)} & 71.5 &73.5& \bf{56.6} &\bf{23.8}\\
    		\hline
	    \end{tabular}
		\end{center}
		
		\label{tab:complt}
	\end{table}
	
	\subsection{Semantic Scene Completion} 
	
	With the semantic guidance from segmentation network, our JS3C-Net can also make significant breakthrough in semantic scene completion (SSC) task. 
	Tab.~\ref{tab:complt} illustrates the results of SSC on SemanticKITTI benchmark, where we compare our JS3C-Net with recent state-of-the-art methods. All methods are implemented with same settings and the detailed implementation and concrete results will be further elaborated in supplementary.
	
	Since semantic scene completion requires to simultaneously predict the occupation status and the semantic label of a voxel, we follow the evaluation protocol of \cite{Song2017Semantic} to compute the precision, recall and IoU for the task of scene completion (SC) while ignoring the semantic label. Meanwhile, the mIoU over the 19 classes is also exploited for the evaluation of semantic scene completion (SSC). As shown in Tab.~\ref{tab:complt}, our JS3C-Net achieves state-of-the-art results on both SC and SSC tasks. Benefit from semantic guidance, joint learning strategy and well-designed interaction module, our SSC module can generate more faithful geometric details. The results of our methods are 6\% higher than previous state-of-the-art TS3D$^3$~\cite{garbade2019two,behley2019semantickitti,liu2018see} for scene completion, which uses segmentation results from DarkNet53Seg~\cite{behley2019semantickitti} as inputs as well. 
	%
	Fig.~\ref{fig:fig5} shows visualization results of semantic scene completion.
	
	\begin{table}
	\small
		\caption{Ablation study on \textit{SemanticKITTI} \textbf{validation set} for semantic segmentation (SS), semantic scene completion (SSC) and scene completion (SC).
		}
		\begin{center}
		\begin{tabular}{c|ccc|c|cc}
			\hline
			Model & JL & UMTL & PVI  & SS & SSC & SC\\
			\hline
			\hline
			A &  &    &
			& 63.1 & 19.4 & 51.1\\
			B & \cmark &  & 
			& 66.1 &22.6 &55.0\\
			C & \cmark & \cmark &
			& 66.4 &23.0 &56.1\\
			D & \cmark & \cmark & \cmark
			& \bf{67.5} &\bf{24.0} &\bf{57.0}\\
			\hline
		\end{tabular}
		\end{center}
		
		\label{tab:abl}
	\end{table}
	
	\begin{figure}[t]
		\centering
		\includegraphics[width=1\columnwidth]{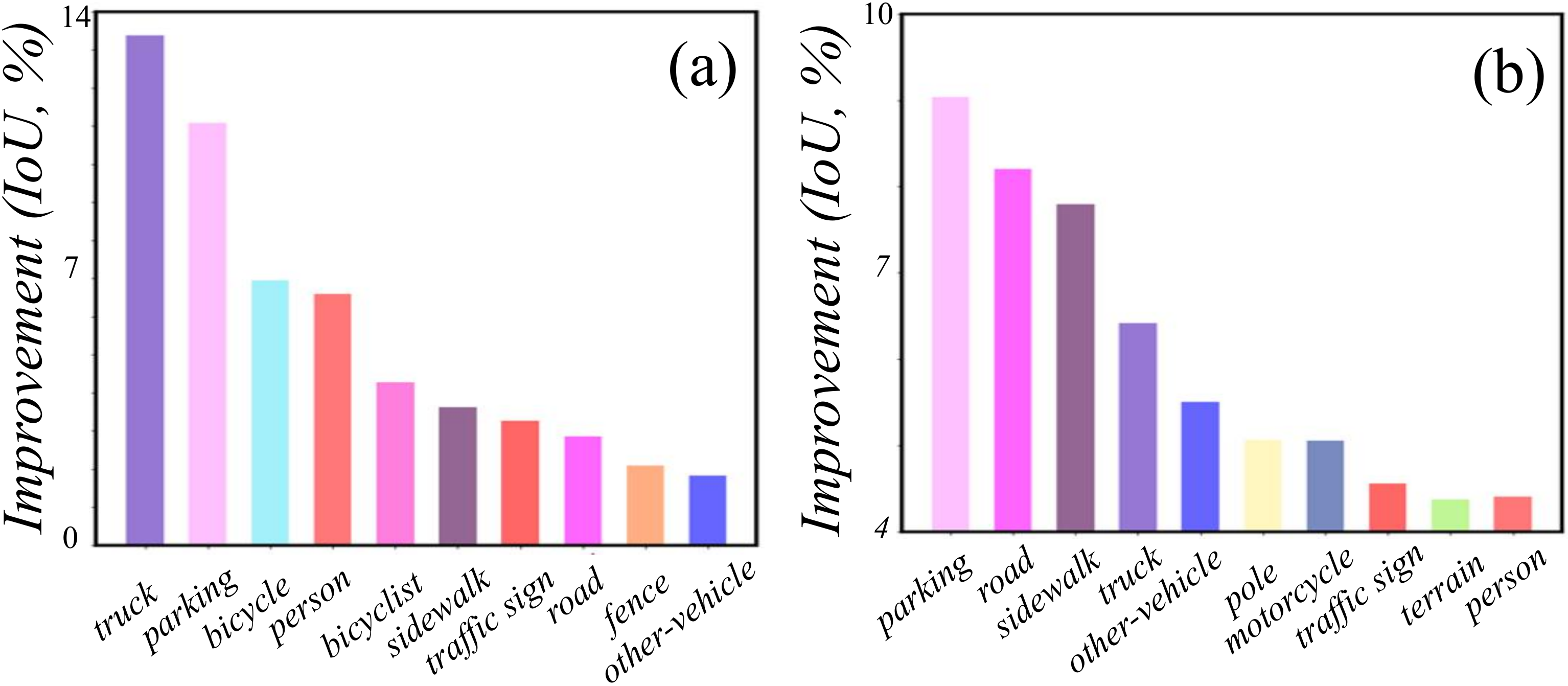}
		
		\caption{Top-10 mIoU gains between JS3C-Net and split-trained single task (SS or SSC) on the \textbf{validation set} of \textit{SemanticKITTI}~\cite{behley2019semantickitti}, where (a) and (b) illustrate SS and SSC respectively.}
		
		\label{fig:fig6}
	\end{figure}
	
	\subsection{Design Analysis}

	\noindent \textbf{Ablation Study.}
    The ablation results are summarized in Tab.~\ref{tab:abl}. Since the limit of submission times, all ablated networks are tested on the validation set of SemanticKITTI dataset~\cite{behley2019semantickitti}. 
	
	The baseline (model A) is set to learn without joint learning, i.e., train two tasks separately. 
	The baseline only gets IoU of 63.1\% on semantic segmentation (SS) and 51.1\% on scene completion (SC). 
	This convincingly confirms the effectiveness of joint learning (JL) in model B (we manually set weights of task losses as 1:0.8), which is significantly improved to 66.1\% for SS and 55.0\% for SC by a large margin. 
	Correspondingly, in model C, uncertainty multi-task loss (UMTL) is used to achieve the optimal trade-off between two tasks automatically.
	As a result, improvements of 0.4\% and 1.1\% on SS and SC are further obtained through UMTL (model C). 
	Then, with PVI module for knowledge fusion, our JS3C-Net achieves the best results on both tasks.
	%

	\noindent \textbf{Mutual Promotion.}
	To further study the reciprocal effects between two tasks, we conducted comparative experiments between the single task (SS or SSC) and the multiple tasks (JS3C-Net). 
	As shown in Fig.~\ref{fig:fig6}, when the JS3C-Net is introduced (i.e., SS and SSC learn jointly), the performances of the two tasks are both largely enhanced, where we show the top 10 mIoU gains in Fig.~\ref{fig:fig6}. 
	It shows that the IoUs of all 19 classes have been improved, especially in the case of small objects, such as trucks, bicycles and persons. 
	The plausible explanation is that, for small objects, their raw point clouds are very sparse and usually lose local details. 
	When SS and SSC learn jointly, SS can take advantage of the contextual shape prior of small objects from SSC, which benefits SS to classify each point more precisely. Therefore, these two tasks can benefit one from the other mutually.

	\noindent \textbf{Complexity Analysis.}
	In this section, we evaluate the overall complexity of JS3C-Net.
	As shown in Tab.~\ref{tab:compl}, our proposed JS3C-Net is much more light-weighted and faster than previous point-based methods (\textbf{1/5} model size and \textbf{1/3} inference time of KPConv).
	More importantly, since the disposable properties of our SSC module and PVI module, JS3C-Net has the same speed with segmentation backbone, which makes it more suitable for the real-time applications.
	
	\begin{table}
	\small
		\caption{\textbf{Complexity Analysis.} Model size and latency for different methods. Here {\underline{underline}} correspond to the post-processing time.
	    Parameters and time in $(\cdot)$ represents the extra operation of SSC and PVI modules, which can be ignored during inference of SS.
		}
		\begin{center}

		\begin{tabular}{lcc}
			\hline
    		& Parameters & Latency \\
    		& (million) & (ms) \\
    		\hline
    		\hline
    		PointNet++&  6.0 & 5900  \\
    		TangentConv&  0.4 & 3000  \\
    		RandLA-Net &  1.2 &  {{256}}+{\underline{624}} \\
    		KPConv &  18.3 &  {{1117}}+{\underline{624}} \\
    		SparseConv &  2.7 & {471} \\
    		\textbf{JS3C-Net (Ours)} &  2.7(+0.4)& \textbf{471}(+107) \\

			\hline
    	\end{tabular}
		\end{center}
		
		\label{tab:compl}
	\end{table}
	
	\section{Conclusion}
	In this work, we propose an single sweep LiDAR point cloud semantic segmentation framework via contextual shape priors from semantic scene completion network, named JS3C-Net. 
	By exploiting some sophisticated pipelines, interactive modules, and reasonable loss function, our JS3C-Net model achieves state-of-the-art results on both semantic segmentation and scene completion tasks, outperforming previous methods by a large margin. We believe that our work can be applied to a wider range of other scenarios in the future, such as indoor point cloud sequence. Meanwhile, our method provides an alternative solution to the comprehension of large-scale LiDAR scenes with severe local details missing. It can improve the performance through contextual shape priors learning and interactive knowledge transferring.
	
	\clearpage
	\newpage
	\section*{Acknowledgments}
    The work was supported in part by the Key Area R\&D Program of Guangdong Province with grant No.2018B030338001, the National Key R\&D Program of China with grant No.2018YFB1800800, NSFC-Youth 61902335, Guangdong Regional Joint Fund-Key Projects 2019B1515120039, Shenzhen Outstanding Talents Training Fund, Shenzhen Institute of Artificial Intelligence and Robotics for Society, Guangdong Research Project No.2017ZT07X152 and CCF-Tencent Open Fund.

	\newpage

	\begin{center}
		{\textit{\LARGE\bf Supplementary Material}}
	\end{center}

	\thispagestyle{empty}
	\setcounter{section}{0}
	\setcounter{figure}{0}
	\setcounter{table}{0}
	\renewcommand\thesection{\Alph{section}}

	In this supplementary material, we first show our experimental details. 
	Besides, we provide additional discussion to further demonstrate the superiority of our model.
	
	\section{Concrete Experimental Design}
	\label{concrete}
	\subsection{Network Architectures}
	
	Our model, JS3C-Net, consists of three parts: a semantic segmentation network, a semantic scene completion (SSC) decoder, and a point-voxel interaction (PVI) module. 
	
    For semantic segmentation stage, we use Submanifold SparseConv~\cite{SparseConv} as our backbone due to its superior performance. 
    The encoder consists of $7$ blocks of 3D sparse convolutions, each of which has two 3D sparse convolutions inside. Going deeper, we gradually increase the number of channels (i.e., $16,32,48,64,80,96,112$). 
    We also apply a 3D sparse pooling operation after each block to reduce the spatial resolution of the feature maps. 
    For the decoder, we use the same structure but in the reverse order and replace the 3D sparse pooling layers with unpooling operations.
    Furthermore, it concatenates the features from the encoder phase on the decoder features at each scale through skip-connection. 
    Three MLPs are used to generate shape embedding for PVI module and C-category prediction.

	The architecture of SSC decoder is already shown in the manuscript. 
	In fact, we use five basic blocks and the channel dimension of each block is identically $32$. 
	As for the point-voxel interaction (PVI) module, graph-based edge learning is implemented by a three-layer MLP with $32$ channels.
	In experiment, we set number of GCN layer $l=1$.
	%

	\subsection{Semantic Segmentation}
	In this section, we illustrate in details the concrete results of semantic segmentation and implementation of compared methods.
	For SemanticKITTI benchmark, we directly compare our results with results on official benchmark.
	For SemanticPOSS dataset (see Tab.\ref{suptab:seg}, we use results of PointNet++ and SequeezeSegV2 in their official paper, and results of RandLA-Net and KPConv are implemented by official codes of correspoding methods.
	
	\subsection{Semantic Scene Completion} 
	In this section, we illustrate in details the concrete results of semantic scene completion and implementation of compared methods, which are already provided by~\cite{behley2019semantickitti}. Tab.~\ref{suptab:complt} depicts the results of semantic scene completion on benchmark~\cite{behley2019semantickitti}.
	Among the methods to be compared, SSCNet~\cite{Song2017Semantic}, EsscNet~\cite{Jiahui2019Efficient}, TS3D~\cite{garbade2019two} directly use sparse point cloud as input, TS3D$^2$ and TS3D$^3$ are multi-stage methods, which use segmentation results of DarkNet53Seg~\cite{behley2019semantickitti} as semantic priors.  
	
	\noindent\textbf{SSCNet.} SSCNet~\cite{Song2017Semantic} is a weak baseline that directly uses processed volumes as input, then several 3D convolution layers with the same structure introduced in their paper are conducted. 
	
	\noindent\textbf{TS3D.} Two Stream (TS3D) approach~\cite{garbade2019two} makes use of the additional information (processed from pre-trained DeepLab\_v2) from the RGB image corresponding to the input laser scan, and then it performs a 4 fold downsampling in a forward process but it renders them incapable of dealing with details of the scene.
	
	\noindent\textbf{TS3D$^2$.} It uses the semantic segmentation results of DarkNet53Seg~\cite{behley2019semantickitti} to enhance the semantic scene completion. However, without a suitable joint learning strategy, the pre-processed feature cannot improve the result of semantic scene completion effectively and significantly. 
	
	\noindent\textbf{TS3D$^3$.} It replaces the backbone of the TS3D$^2$ with SATNet~\cite{liu2018see} without downsampling, and divide each whole volume into six equal parts for more fine-grained prediction. 
	
	\noindent\textbf{EsscNet.} We also compare our method with  EsscNet~\cite{Jiahui2019Efficient}, which uses spatial group convolution (SGC) for memory saving. It uses the entire scene as input. However, due to the lack of priors, it still cannot perform very well. Also, the large amount of parameters shown in the manuscript makes it unable to meet the requirement for real-time applications.
	
	
		\begin{table}
	\small
	\caption{The results of different joint learning strategies on \textit{SemanticKITTI} validation set. SS,  SC and SSC mean semantic segmentation, scene completion and semantic scene completion, respectively. JL means learn two tasks jointly.}

	\begin{center}
		\begin{tabular}{c|cccc|c|cc}
		\hline
		Model & SS & SC & SSC & JL  & SS & SSC & SC\\
		\hline
		\hline
		A & \cmark &   &  &
		& 63.1 & 19.4 & 51.1\\
		B & \cmark & \cmark &  
		& \cmark & 62.5 & - & 55.7\\
		C & \cmark & & \cmark  
		& \cmark & \bf{67.5} &\bf{24.0} &\bf{57.0}\\
		\hline
	    \end{tabular}
		\end{center}
		
		\label{tab:sc}
	\end{table}
	
	    \begin{table}
	\small

	\begin{center}
	\caption{Effect of moving objects on \textit{SemanticKITTI} and \textit{SemanticPOSS} validation set for semantic segmentation, where w and w/o mean 'with' and 'without'.}
	\begin{tabular}{lcc}
		\hline
		& SemanticKITTI & SemanticPOSS \\
		\hline
		\hline
		w moving objects &  67.5 & 60.2 \\
		w/o moving objects &  {67.7}  &  60.2\\
		\hline
	\end{tabular}
		\label{tab:mov}
	\end{center}

	\end{table}

	\begin{table*}[]
		\small
		\caption{Semantic segmentation results on the \textit{SemanticPOSS} dataset.}
		\begin{center}
			\resizebox{\textwidth}{!}{%
				\begin{tabular}{l|ccccccccccc|c}
					\hline
					
					Method &
					people&
					rider&
				    car&
					traffic sign&
					trunk&
					plants&
					pole&
					fence&
					building&
					bike&
					road&
					mIoU \\
					\hline
					\hline
					
					PointNet++~\cite{qi2017pointnet,qi2017pointnet++} &20.8 &0.1 &8.9 &21.8 &4.0 &51.2 &3.2 &6.0 &42.7 &0.1 &62.2 &20.1 \\
					SequeezeSegV2~\cite{wu2019squeezesegv2} &18.4 &11.2 &34.9 &11.0 &15.8 &56.3 &4.5 &25.5 &47.0 & 32.4 &71.3 &29.8\\
				
				
				
				
					RandLA-Net~\cite{hu2019randla}& 69.2  &26.7 &77.2 &24.3 &27.3 &73.4 &30.9 &51.4 &82.5 &43.9 &81.2 &53.5\\
					KPConv~\cite{Thomas_2019_ICCV}&77.3 &29.4 &78.2 &23.0 &28.1  &71.6 &32.4 &51.5 &81.8 &53.2 &80.6 &55.2 \\
					
					SparseConv (Baseline) & 76.3 &30.5 &80.8 &28.3 &29.1 &74.9 &39.8 &51.8 &83.7 &53.5 &80.5 &57.2\\
					\hline
					\bf{JS3C-Net(Ours)} & \bf{80.0} &\bf{39.1} &\bf{83.2} &\bf{28.9} &\bf{31.6} &\bf{76.1} &\bf{44.7} &\bf{54.0} &\bf{83.9} &\bf{59.8} &\bf{81.7} &\bf{60.2}\\
					\hline
			\end{tabular}}
		\end{center}
		\label{suptab:seg}
	\end{table*}

	\section{Additional Discussion}
	\label{discussion}
	\subsection{Why Use Semantic Scene Completion?}
	In this section, we further discuss why our model should use semantic scene completion as the object of joint learning. 
	Note that semantic segmentation (SS) and semantic scene completion (SSC) both include semantic labeling process, which seems to be redundant in these two tasks. 
	Therefore, we conduct an experiment that uses scene completion (SC) instead of semantic scene completion (SSC) for the joint learning with semantic segmentation (SS).

	As shown in Tab.~\ref{tab:sc}, when using scene completion and semantic segmentation for joint learning (model B), although it can somehow improve the result of scene completion, the segmentation results become worse than directly conducting segmentation network (model A). This proves that directly adding scene completion will introduce more bias into segmentation network.
	
	The explanation for such results can be summarized as follows: (1) Due to the cascaded architecture of our model, the results of SS and SSC actually have spatial alignment, which allows our SSC decoder to focus more on completing the shapes of each category rather than the entire scene. This setting allows the process of back propagation to bring shape prior to each category of semantic segmentation with less noise. 
	(2) PVI module uses both spatial coordinates and features information to construct a learnable weight of graph edges, where features from SS and SSC are mainly used to compare the differences in feature space.
	Therefore, it can further enhance the distinction of features in segmentation network through using 20 categories to predict the SSC with strong semantic priors instead of simply predicting whether a certain voxel has an object or not.

	\subsection{Do Moving Objects Affect the Results?}
	
\begin{figure*}[t]
        \centering
        \includegraphics[width=\columnwidth]{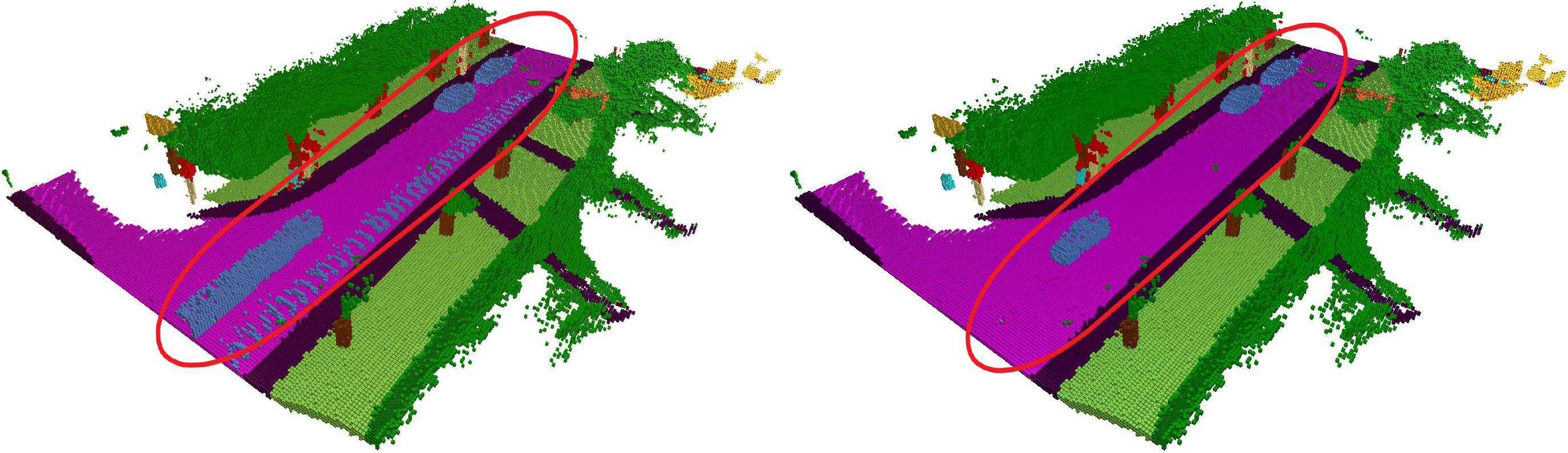}
        \caption{The selected example of ground truths for SSC, where (left) and (right) illustrate reconstructed results with and without moving objects.} 
        \label{fig:moving}
    \end{figure*}

	As shown in the red circle of Fig.~\ref{fig:moving} (left), when using multiple frames to generate an entire dense scene volume, moving objects will be inevitably reconstructed into a long bar, which may introduce bias to the semantic segmentation. 
	Therefore, to further test whether these moving objects will affect segmentation results, we design an experiment as follows. Firstly, we normally reconstruct moving objects, as shown in Fig.~\ref{fig:moving} (left). Then, as shown in Fig.~\ref{fig:moving} (right), we use the annotations of moving objects in both two datasets, and only use $5$ frames to reconstruct each moving object (instead of $70$ frames).
	
	The quantitative results are shown in Tab.~\ref{tab:mov}. It can be seen that, when we remove moving objects in semantic scene completion, there is only $0.2\%$ improvement in semantic segmentation.  This is because moving objects only account for a small proportion in the datasets. 
	Finally, since there are also moving objects in SSC benchmark of~\cite{behley2019semantickitti}, we keep moving objects in our training data in order to make a fair comparison with other methods.
	
	\subsection{Can We Only Consider Historical Frames?}
	To further illustrate the advantages of using semantic scene completion, we also compare the results of only consider historical frames.
	Here we consider the method proposed in SpSequenceNet~\cite{shi2020spsequencenet} to use attention mechanism and feature aggregation from historical frames.
	
	Tab.~\ref{tab:his} shows the results of the correlative ablation study.
	Since results of SpSequenceNet on SemanticKITTI is much lower than ours, we re-implement their framework with our baseline.
	Experiment results show that only using attention or feature aggregation with previous frames cannot better learn the complete shape priors from LiDAR sequences.
	Our results of only using historical frames to generate ground truth is still higher than their results.
	Note that only using historical frames will ignore the shape details of newly incoming objects in each frame, thus hampers the results.
	When we consider future frames of LiDAR sequence, our results will be further improved. 
	
	Furthermore, their method inevitably introduces more computational burden to segmentation, while our auxiliary components are fully discardable (see latency in table).
	
	\begin{table}
	\small
	\caption{The results on \textit{SemanticKITTI} validation set.}

	\begin{center}

	\begin{tabular}{lcc}
		\hline
		Ablation & Latency (ms)& mIoU \\
		\hline
		\hline
		SparseConv (baseline) & 471 & 63.1  \\
		SparseConv + SpSequenceNet & 972 & {64.3}  \\
		JS3C-Net (historical) & 471 & {66.6}  \\
		JS3C-Net (historical+future) & 471 & \textbf{67.5}  \\
		\hline
	\end{tabular}
	\end{center}
		
	\label{tab:his}
	\end{table}
	
		\begin{table*}[ht]
		\small
		\renewcommand\tabcolsep{1.5pt} 
		\caption{Semantic scene completion results on the \textit{SemanticKITTI} benchmark. Only the recent published approaches are compared.}
		\begin{center}
			\resizebox{\textwidth}{!}{%
				\begin{tabular}{lccc|ccccccccccccccccccc|c}
					\hline
					
					& \multicolumn{3}{c|}{Scene Completion}
					&\multicolumn{19}{c|}{Semantic Scene Completion} &\\
					Method
					&\rotatebox{90}{precision}&
					\rotatebox{90}{recall}&
					\rotatebox{90}{IoU}&	
					\rotatebox{90}{road}&
					\rotatebox{90}{sidewalk}&
					\rotatebox{90}{parking}&
					\rotatebox{90}{other-ground}&
					\rotatebox{90}{building}&
					\rotatebox{90}{car}&
					\rotatebox{90}{truck}&
					\rotatebox{90}{bicycle}&
					\rotatebox{90}{motorcycle}&
					\rotatebox{90}{other-vehicle}&
					\rotatebox{90}{vegetation}&
					\rotatebox{90}{trunk}&
					\rotatebox{90}{terrain}&
					\rotatebox{90}{person}&
					\rotatebox{90}{bicyclist}&
					\rotatebox{90}{motorcyclist}&
					\rotatebox{90}{fence}&
					\rotatebox{90}{pole}&
					\rotatebox{90}{traffic sign}&
					\rotatebox{90}{mIoU} \\
					\hline
					\hline
					
					SSCNet~\cite{Song2017Semantic} &31.7 &83.4 &29.8 &27.6 &17.0 &15.6 &6.0 &20.9 &10.4 &1.8 &0.0 &0.0 &0.1 &25.8 &11.9 &18.2 &0.0 &0.0 &0.0 &14.4 &7.9 &3.7 &9.5\\
					TS3D~\cite{garbade2019two} &31.6 &84.2 &29.8 &28.0 &17.0 &15.7 &4.9 &23.2 &10.7 &2.4 &0.0 &0.0 &0.2 &24.7 &12.5 &18.3 &0.0 &0.1 &0.0 &13.2 &7.0 &3.5 &9.5\\
					TS3D$^2$~\cite{garbade2019two,behley2019semantickitti} &25.9 &\bf{88.3} &25.0 &27.5 &18.5 &18.9 &{6.6} &22.1 &8.0 &2.2 &0.1 &0.0 &4.0 &19.5 &12.9 &20.2 &2.3 &0.6 &0.0 &15.8 &7.6 &6.7 &10.2\\
					EsscNet~\cite{Jiahui2019Efficient} &62.6 &55.6 &41.8 &43.8	&28.1	&26.9	&10.3 &29.8	&26.4 &5.0	&0.3	&5.4	&9.1	&35.8	&\bf{20.1}	&28.7	&2.9	&2.7	&0.1	&23.3	&16.4	&\bf{16.7} &17.5\\
					TS3D$^3$~\cite{garbade2019two,behley2019semantickitti,liu2018see} &\bf{80.5} &57.7 &50.6 &62.2 &31.6 &23.3 &6.5 &34.1 &30.7 &4.9 &0.0 &0.0 &0.1 &40.1 &21.9 &33.1 &0.0 &0.0 &0.0 &24.1 &16.9 &6.9 &17.7\\
					\hline

					\textbf{JS3C-Net (Ours)} & 70.2 &74.5& \bf{56.6} &\bf{64.7} &\bf{39.9} &\bf{34.9} &\bf{14.1} &\bf{39.4} &\bf{33.3} &\bf{7.2} &\bf{14.4} &\bf{8.8} &\bf{12.7} &\bf{43.1} &19.6 &\bf{40.5} &\bf{8.0} &\bf{5.1} &\bf{0.4} &\bf{30.4} &\bf{18.9} &15.9 &\bf{23.8} \\
					\hline
			\end{tabular}}
		\end{center}
		\label{suptab:complt}
	\end{table*}

	  \begin{figure*}[ht]
		
		\noindent\includegraphics[width=\columnwidth]{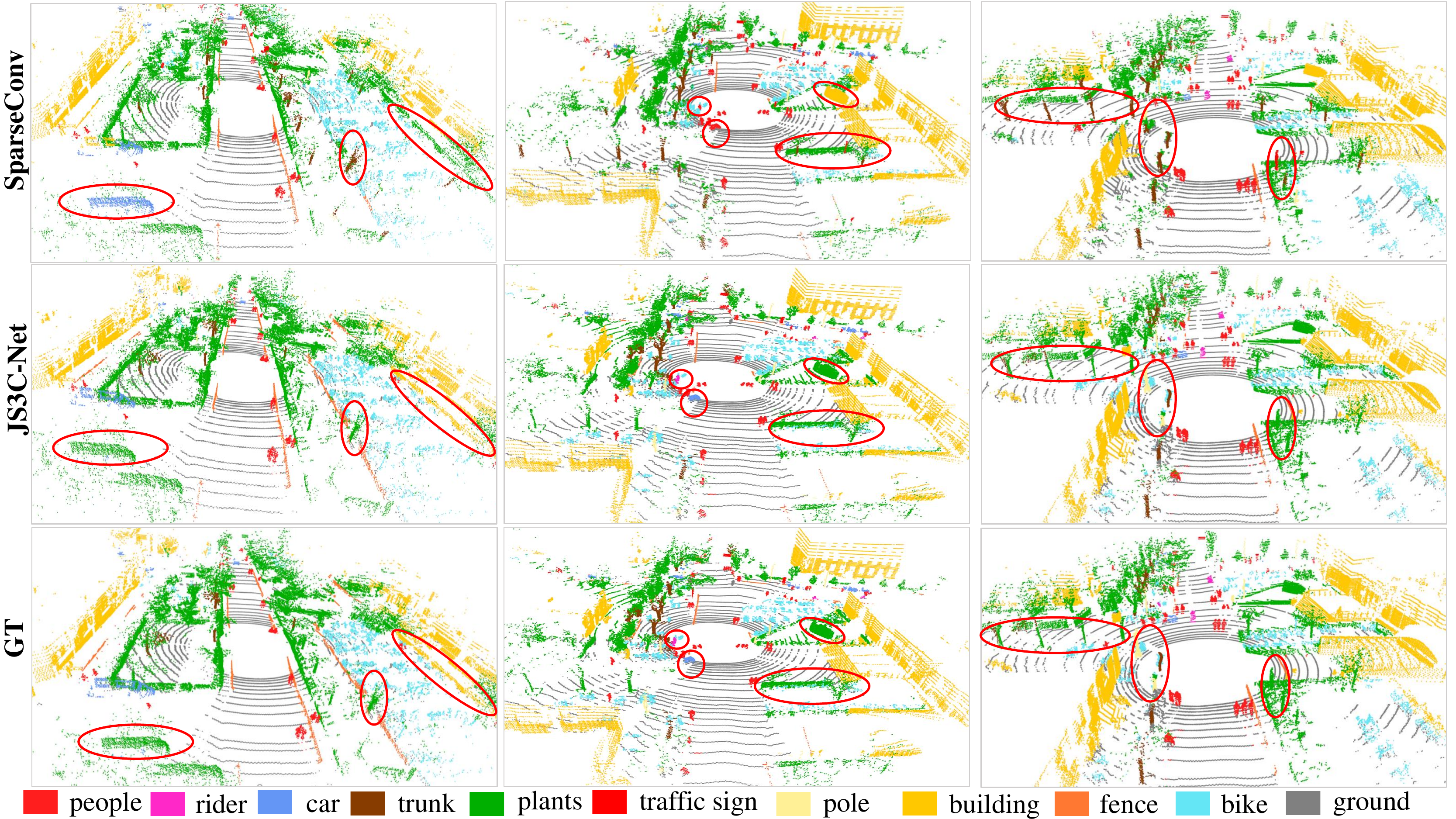}
		
		\caption{{The visualization results on \textit{SemanticPOSS} dataset.}}
		\label{supfig:fig4}

	\end{figure*}

    \section{More Visualization}

    In Fig.~\ref{supfig:fig4}, we show more cases on SemanticPOSS dataset.    
\end{document}